\pdfoutput=1

\documentclass[11pt]{article}

\usepackage{emnlp2022}

\usepackage{times}
\usepackage{latexsym}

\usepackage[T1]{fontenc}

\usepackage[utf8]{inputenc}

\usepackage{microtype}

\usepackage{inconsolata}

\usepackage{graphicx,subfigure,multirow}
\usepackage{amsmath, bm, amsfonts}
\usepackage{amsthm}
\usepackage{booktabs}
\usepackage{algorithm, algpseudocode}
\usepackage{color}

\usepackage{todonotes}
\usepackage{cleveref}
\newcommand{\softmax}{\mathrm{softmax}}
\newcommand{\LeakyReLU}{\mathrm{LeakyReLU}}
\newcommand{\layernorm}{\mathrm{LN}}
\crefname{section}{§}{§§}

\title{Modeling Semantic Composition with Syntactic \\Hypergraph for Video Question Answering}

\author{Zenan Xu\thanks{\ \ \ Equal Contribution}~~, Wanjun Zhong$^*$, Qinliang Su$^*$\thanks{~~~Corresponding to suqliang@mail.sysu.edu.cn}~, Zijing Ou and Fuwei Zhang}

\begin{document}
\maketitle
\begin{abstract}
	A key challenge in video question answering is how to realize the cross-modal semantic alignment between textual concepts and corresponding visual objects. Existing methods mostly seek to align the word representations with the video regions. However, word representations are often not able to convey a complete description of textual concepts, which are in general described by the compositions of certain words. To address this issue, we propose to first build a syntactic dependency tree for each question with an off-the-shelf tool and use it to guide the extraction of meaningful word compositions. Based on the extracted compositions, a hypergraph is further built by viewing the words as nodes and the compositions as hyperedges. Hypergraph convolutional networks (HCN) are then employed to learn the initial representations of word compositions. Afterwards, an optimal transport based method is proposed to perform cross-modal semantic alignment for the textual and visual semantic space. To reflect the cross-modal influences, the cross-modal information is incorporated into the initial representations, leading to a model named cross-modality-aware syntactic HCN. Experimental results on three benchmarks show that our method outperforms all strong baselines. Further analyses demonstrate the effectiveness of each component, and show that our model is good at modeling different levels of semantic compositions and filtering out irrelevant information.
\end{abstract}

\section{Introduction}
Video question answering (VideoQA) requires systems to understand the visual information and infer an answer for a natural language question from it. It has emerged as an important task with notable development towards bridging the gap between computer vision and natural language. VideoQA is challenging as it needs to understand the complex cross-modal relation between natural language question and video. 

\begin{figure}[!t]
	\centering
	\includegraphics[scale=0.38]{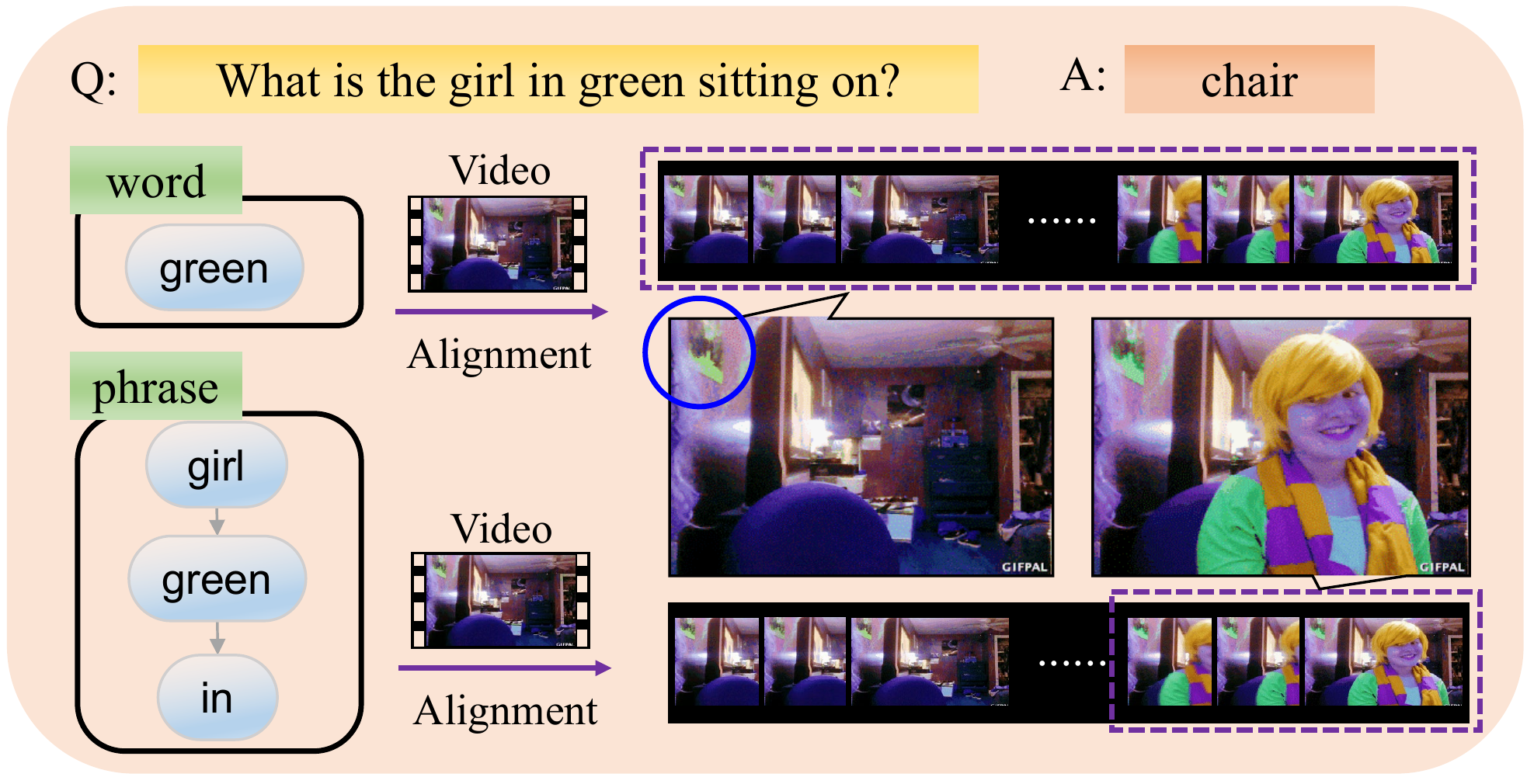}
	\caption{Different levels of semantic composition would be aligned to different frames.}
	\label{fig:introduction}
\end{figure}

To capture the visual-language relation, some works have been proposed to utilize bilinear pooling operation or spatial-temporal attention mechanism to allign the video and textual features~\cite{Jang2019VideoQA,Seo2021AttendWY}. Some methods also proposed to use the co-attention mechanism~\cite{Jiang2020ReasoningWH,Li2021RelationawareHA} to align multi-modal features, or use memory-augmented RNN~\cite{Yin2020MemoryAD} or graph memory mechanism~\cite{liu2021hair} to perform relational reasoning in VideoQA. Recently, DualVGR~\cite{wang2021dualvgr} devises a graph-based reasoning unit and performs a word-level attention to obtain the question-related video features. In these methods, a cross-modal alignment between textual concepts and visual objects is attempted to be found, which, however, is mostly done at the word level, {\it i.e.}, aligning the representations of words with visual objects/frames. However, a word representation (even the contextual representation obtained from LSTM) is often not able to convey a complete description of a textual concept, which plays an essential role in video question answering. For instance, to answer the question ``{\em What is the girl in green sitting on?}" as seen in Fig.~\ref{fig:introduction}, the model need to understand the textual concepts of ``{\em girl in green}'', ``{\em girl sitting on}'' etc., then align them with the corresponding video regions. But if the alignment is done at the word level, the word ``{\em green}'' will be aligned with all green objects in the video, $e.g.$, the green painting on the wall. Obviously, this is not the real intent of this question. In a question, there are often many textual concepts at different semantic levels, such as ``{\em green}'', ``{\em girl in green}'' and ``{\em girl in green sitting on}'' etc.  Therefore, in the task of VideoQA, it is extremely beneficial to identify meaningful textual concepts at different semantic levels. A textual concept is generally described by a compositions of word that are not necessary to be adjacent, thus we cannot simply use the chunks of consecutive words to represent them. 

\begin{figure}[!t]
	\centering
	\includegraphics[scale=0.4]{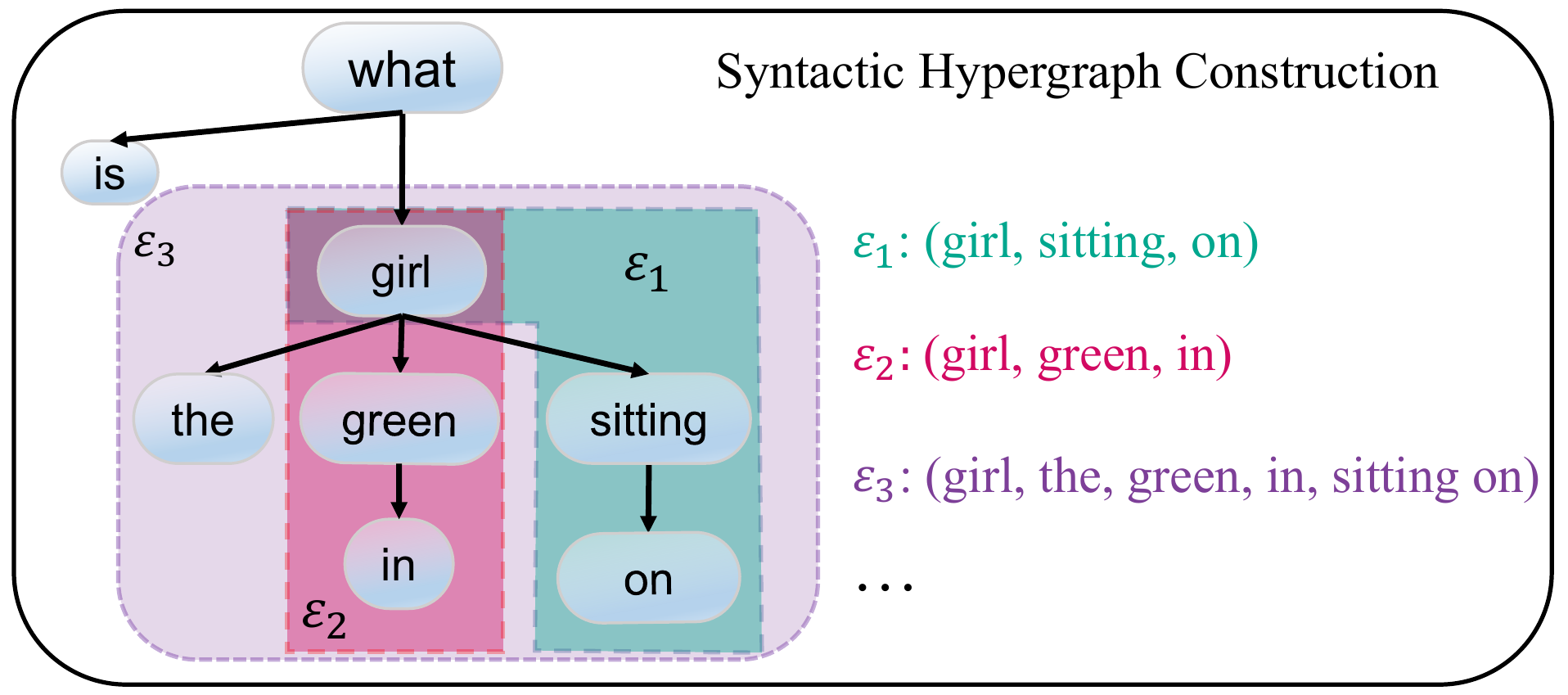}
	\caption{Illustration of syntactic hypergraph construction.}
	\label{fig:methodology_1}
\end{figure}

To obtain meaningful compositions of words, we find that this problem is closely related to the syntactic dependency tree~\cite{tai2015improved}, which describes the dependence structure of words in a sentence. We notice that every subtree can be approximately used to represent a meaningful composition of words that represents a textual concept, as illustrated in Fig. \ref{fig:methodology_1}. Moreover, different-level textual concepts can be effectively captured by the subtrees of different orders. Thus, by building a syntactic dependency tree for questions with an off-the-shelf tool, we are able to obtain a set of compositions of words that representing different textual concepts. A hypergraph is further built by viewing the words as nodes and the compositions as hyperedges. Hypergraph convolutional networks (HCN) are then employed to learn the initial representations of these compositions (textual concepts). Given the initial representations, an optimal transport (OT) based alignment mechanism is developed to align the textual concepts and visual objects, which has been shown to be better at producing more accurate and sparser alignment than methods of directly using dot-product to compute the similarities ($e.g.,$ cosine similarity) \cite{Niculae2017ARF,chen2020improving}. With the cross-modal alignment, we further propose to update the initial composition and video representations by incorporating the relevant cross-modal information into them. To demonstrate the effectiveness of our approach, we compare our approach with competitive baselines on three benchmark VideoQA datasets, including TGIF-QA~\cite{Jang2017TGIFQATS}, MSVD-QA~\cite{Xu2017VideoQA}, and MSRVTT-QA~\cite{Xu2016MSRVTTAL,Xu2017VideoQA}. The experimental results show that \textit{SCAN}\footnote{SCAN stands for \textbf{S}emantic \textbf{C}omposition and \textbf{A}lignment with Cross-Modality-Aware Syntactic Hypergraph Convolution \textbf{N}etwork.} outperforms all strong baselines and further analyses verify the validity of each component. Qualitative analysis further illustrates that our methods performs better in matching the video information based on semantic composition of text, and also in alleviating noise. 

\section{Related Work}
The VideoQA task requires machine to understand the visual-language correlation~\cite{Jang2019VideoQA,Le2020HierarchicalCR,Zhang2021FusingTD,wang2021dualvgr,Guo2021AUQ}. To do this, MDAM~\cite{Kim2018MultimodalDA} and PSAC~\cite{Li2019BeyondRP} proposes to adopt the self-attention based approaches to learn the correlation between each frame and question. To enhance the frame-question correlation, L-GCN~\cite{huang2020location} and HAIR~\cite{liu2021hair} first extracts the object information from each frame and integrate both object-level and frame-level information to enhance the frame-question correlation. Some researches have attempted to capture more fine-grained visual-language correlation. MASN~\cite{Seo2021AttendWY} introduce frame-level and clip-level modules to simultaneously model different-level correlation between visual information and question. RHA~\cite{Li2021RelationawareHA} proposed to use hierarchical attention network to further model the video subtitle-question correlation. There are also researches that adopt the memory-augmented approaches to capture the correlation~\cite{Fan2019HeterogeneousME,Yin2020MemoryAD}. Although these works have effectively capture the visual-language correlation, they ignore the syntactic compositional semantics of fine-grained concepts in the question, our \textit{SCAN} bridges this gap by introducing a syntactic hypergraph and performing visual-aware hypergraph convolution.

\begin{figure*}[!t]
	\centering
	\includegraphics[scale=0.5]{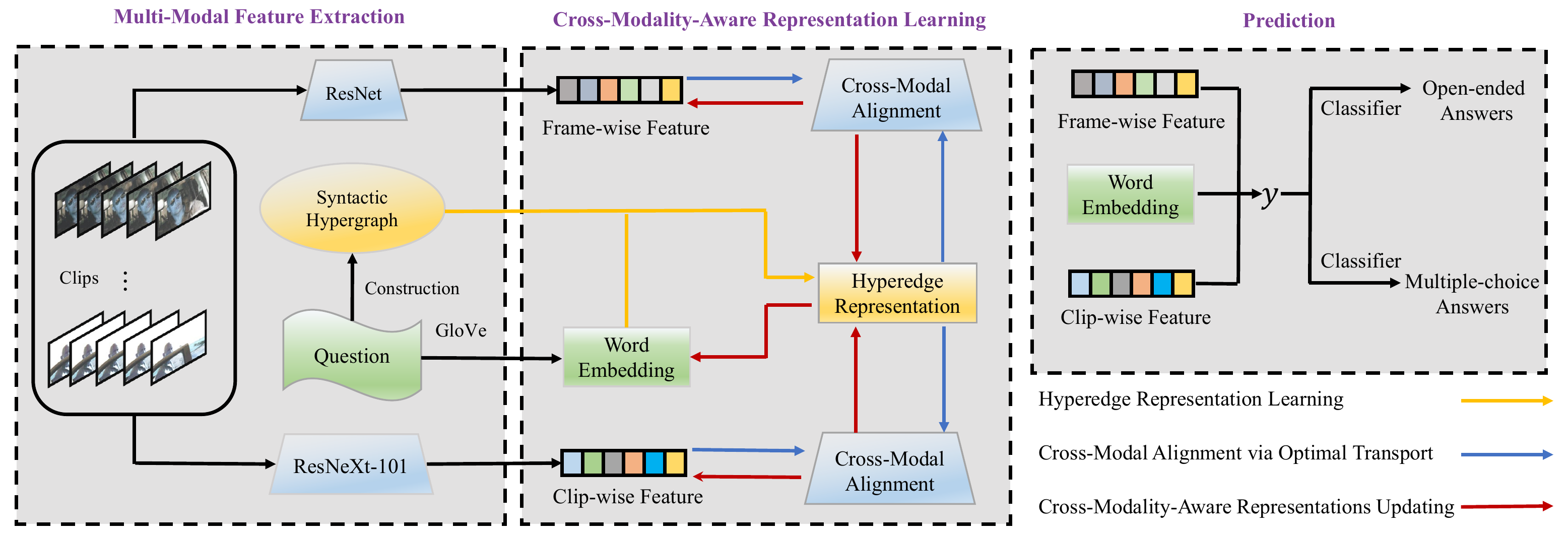}
	\caption{An overview of our proposed \textit{SCAN}.}
	\label{fig:methodology_2}
\end{figure*}

\section{Methodology}
In this section, we first briefly introduce the VideoQA task with basic notation, and then describe the definition and construction of the syntactic hypergraph . Based on the syntactic hypergraph, we present our cross-modality-aware syntactic hypergraph convolutional network to model the multi-modal interaction between the question and the video, where the optimal transport is employed for the alignment. The overall framework of our proposed \textit{SCAN} is shown in Fig.~\ref{fig:methodology_2}.

\subsection{Preliminaries} \label{sec:Preliminaries}
\paragraph{Task Definition}
Given a video $V$ and a question $Q$, the VideoQA task requires the system to find the answer $a\in\mathcal{A}$ with maximum probability $p(a|V, Q, \theta)$, where $\theta$ denotes the model parameters. Answers for VideoQA task are usually organized in two forms, $i.e.$, open-ended form and multiple-choice form. The open-ended answer is represented as a free-form text, while the answer of each multiple-choice question comes from a set with fixed number of answer candidates. 

\paragraph{Multi-modal Features}
Following previous works for VideoQA~\cite{Park2021BridgeTA} in which several consecutive frames in a video will be combined into one clip, we will first divide a video into several clips of the same length. Then we separately employ the pre-trained ResNet~\cite{He2016DeepRL} and ResNeXt-101~\cite{Hara2018CanS3} model on each frame and clip to extract the frame-wise appearance feature matrix $\bm{F}\in\mathbb{R}^{N_f\times d_v}$ and clip-wise motion feature matrix $\bm{M}\in\mathbb{R}^{N_c\times d_v}$, where $d_v$ is the dimension of video feature (usually to be 2048), and $N_f$ and $N_c$ denote the number of frames and clips, respectively. For question $Q$, we follow the previous work~\cite{Jiang2020ReasoningWH} to represent each word with the pre-trained GloVe~\cite{Pennington2014GloVeGV} word embedding and obtain question embedding matrix $\bm{Q}\in\mathbb{R}^{N_w\times d_w}$, where $N_w$ is the number of word in a question, and $d_w$ denotes the dimension of word embedding (usually to be 300).

\subsection{Syntactic Hypergraph} \label{sec:Hypergraph Construction}
In this part, we first introduce the definition of hypergraph and the advantage of modeling semantic compositons of words with hypergraph. Then, we describe how to construct hypergraph under the guidance of syntactic dependency tree.
\paragraph{Hypergraph Definition}
Let $\mathcal{G}=\{\mathcal{V},\mathcal{E}\}$ denote a hypergraph, where $\mathcal{V}$ is a set containing $N_v$ vertices and $\mathcal{E}$ is a set containing $N_e$ hyperedges. Each hyperedge $\varepsilon\in\mathcal{E}$ denotes a set containing any numbers of vertices. The hypergraph can be represented by an incidence matrix $\bm{H}\in\mathbb{R}^{N_v\times N_{e}}$ where $\bm{H}_{i,\varepsilon}$ = 1 if the hyperedge $\varepsilon\in\mathcal{E}$ contains the vertex $v_i\in\mathcal{V}$, otherwise 0. 	For example, if there are vertices $\mathcal{V}=\{``{green}",``{girl}",``{sitting}"\}$, one possible hyperedge could be the set $\varepsilon=\{{``{green}",``{girl}"}\}$.
We represent the semantic compositions of words with the hypergraph because the characteristics of hyperedge perfectly fit our assumption that a textual concept is represented by a set-like semantic composition of words. Therefore, it provides us with flexibility and capability to model complex interactions of words in the view of semantic composition.
\paragraph{Syntactic Hypergraph Construction}
Since the semantic composition phenomenon in nature language is usually related to the syntactic properties~\cite{tai2015improved}, we take the syntactic information ($i.e.$ dependency syntax tree) as guidance to model the semantic composition of a question. Specifically, we apply off-the-shelf Stanza\footnote{https://github.com/stanfordnlp/stanza} toolkit on a given question to automatically generate the corresponding syntax tree, which represents the hierarchical syntactic relations of words. As the example shown in Fig.\ref{fig:methodology_1}, each leaf node in the syntax tree is a word, then each subtree can be viewed as a case of semantic composition of corresponding words. Therefore, we define each hyperedge as the set of words in each subtree. We travel the syntax tree in a hierarchical bottom-up manner to find all the subtrees of the syntax tree, and finally build all hyperedges with those subtrees. 

Specifically, we describe the process of subtree generation method as follows. The algorithm begins by taking each leaf node as a subtree. Then we generate more subtrees by recursively adding higher-order branch nodes to the initial trees in a bottom-up manner. In each step, for each branch node, we add it to all its connected subtrees to generate more trees. Take the branch node $``{girl}"$ in Fig.~\ref{fig:methodology_1} as an example, we add $``{girl}"$ to two connected subtrees $\{{``{in}",``{green}"}\}$ and $\{{``{sitting}",``{on}"}\}$ to generate two new subtrees $\{{``{girl}",``{in}",``{green}"}\}$ and $\{{``{girl}",``{sitting}",``{on}"}\}$. Due to space limit, the pseudo algorithm of our subtree generation is given in Appendix.

\subsection{Cross-Modality-Aware Syntactic Hypergraph Convolutional Network} \label{sec:Cross-Modality Hypergraph Convolutional Network}
With the syntactic hypergraph, we now propose a novel cross-modality-aware syntactic hypergraph convolutional network to incorporate cross-modal information into the initial representations, in contrast to the vanilla hypergraph convolutional network that is initially designed to model the information propagation between nodes with hyperedge as the bridge \cite{feng2019hypergraph}.
The model mainly includes three steps: 1) \textbf{Initial hyperedge representation learning}: learning the initial hyperedge (composition) representations from node (word) representations; 2) \textbf{Cross-modal alignment}: finding cross-modal alignment between textual concepts and visual objects; 
3) \textbf{Cross-modality-aware representations updating}: Updating the representations of hyperedges and videos by incorporating the cross-modal information into them.


\paragraph{Initial Hyperedge Representation Learning} 
The first step is to gather node representations to produce the initial hyperedge representation to model the semantic composition process of words. We take contextual features of question words $\bm{Q}$ (Sec.~\ref{sec:Preliminaries}) to initialize the node representations. 
Then, we gather the node representations to produce the corresponding hyperedge representations as:
\begin{equation} \label{equ:n2e}
\bm{X} = \bm{D}^{-1}_e\bm{H}^T\bm{Q}\bm{W},
\end{equation}
where $\bm{H}$ is the 0-1 incidence matrix representing whether a node is connected by a hyperedge (Sec.~\ref{sec:Preliminaries}), and
$\bm{X}\in\mathbb{R}^{N_s\times d_w}$ contains the representation of all $N_s$ hyperedges, $\bm{W}\in\mathbb{R}^{d_w\times d_w}$ is a weight matrix, and $\bm{D}_e$ is the diagonal matrix denoting the degree of the hyperedge, which is defined as the number of the node connected by a hyperedge. It can be seen that, the multiplication operation $\bm{H}^T\bm{Q}$ in \eqref{equ:n2e} performs the information gathering of words in a hyperedge. 
Since each hyperedge aggregates information from a set of nodes that probably represents a textual concept, the question-video alignment problem is then approximately reduced to the matching problem between hyperedge representations and video features.

\paragraph{Cross-Modal Alignment via Optimal Transport} 
\!Given the initial hyperedge representations $\bm{X}$, we now present how to align the textual concepts and video frames. What we want to obtain is an alignment matrix $\bm{G}_{xf} \in\mathbb{R}^{N_s\times N_f}$, whose $(i, j)$-th element can reflect the degree of alignment between the $i$-th textual concept and the $j$-th frame. In this paper, inspired by recent success of optimal transport (OT) in the alignment of sole textual and visual space \cite{chen2020improving}, we propose to apply it to align the cross-modal spaces, {\it i.e.}, the textual and visual semantic spaces. Specifically, by viewing the hyperedge representations $\{\bm{x}_i\}_{i=1}^{N_s}$ and video frame representations $\{\bm{f}_j\}_{j=1}^{N_f}$ as two empirical probability distributions, we can define an optimal transport plan $\bm{\pi}^* \in\mathbb{R}^{N_s\times N_f}_+$ that will transport the features from textual concept space to video frame space with minimum transportation cost. Mathematically, the optimal transport plan is obtained by solving the following optimization problem 
\begin{equation} \label{equ:ot}
\bm{\pi}^* = \mathop{\arg\min}_{\bm{\pi}\in\Pi_\mathcal{I}}\left\lbrace\sum_{ij}{\pi}_{ij}c(\bm{x}_i,\bm{f}_j)\right\rbrace,
\end{equation}
where $\Pi_\mathcal{I}$ denotes the set of all feasible transport plans, which is composed of all  $N_s\times N_f$ non-negative matrices whose elements are summed to be 1; and the cost function $c(\cdot, \cdot)$ is defined as
\begin{equation} \label{OP_cost}
c(\bm{x}_i, \bm{f}_j) = 1 - \frac{\bm{x}_i\bm{T}_x(\bm{f}_j\bm{T}_f)^T}{\Vert\bm{x}_i\bm{T}_x\Vert\cdot\Vert\bm{f}_i\bm{T}_f\Vert};
\end{equation}
and $\bm{T}_x\in\mathbb{R}^{d_w\times d}$ and $\bm{T}_f\in\mathbb{R}^{d_v\times d}$ are used to transform the textual and visual features into the same semantic space. It can be seen that if the $i$-th textual concept and $j$-th frame are semantically relevant, the cost $c(\bm{x}_i, \bm{f}_j)$ will be small, otherwise it will be large.  Thus, if the $i$-th hyperedge is closely relevant to the $j$-th frame, then $\pi_{ij}^*$ will be assigned a large value, otherwise a value close to $0$ will be assigned. Therefore, in this paper, the optimal transport plan $\bm{\pi}^*$ is directly used as the alignment weight matrix, that is, 
\begin{equation}
\bm{G}_{xf} = \bm{\pi}^*.
\end{equation}
It is shown in our experiments that this method tends to produce a sparser alignment matrix $\bm{G}_{xf}$ than the method of directly using dot-product to compute the similarity, and also leads to better results. Generally, the optimization problem \eqref{equ:ot} can not be solved exactly. In this paper, an off-the-shelf differentiable approximate method proposed in~\cite{xie2020fast} is borrowed to obtain the optimal matrix $\bm{\pi}^*$ approximately, where the concrete algorithm is presented in the Appendix. In a similar way, an alignment matrix $\bm{G}_{xm}$ that reflects the alignment degree between hyperedge representations $\bm{X}$ and clip features $\bm{M}$ can also be obtained.

\paragraph{Cross-Modality-Aware Representations Updating} Given the alignment matrices $\bm{G}_{xm}$ and $\bm{G}_{xf}$, we now leverage them to improve the representations of hyperedges and video by incorporating the relevant cross-modal information into the initial representations. Specifically, we propose to compute the influence from video to hyperedge $\bm{X}_{v\rightarrow x}\in\mathbb{R}^{N_s\times d}$ and the influence from hyperedges to frame $\bm{F}_{x\rightarrow f}\in\mathbb{R}^{N_f\times d}$ as
\begin{equation}
\begin{aligned}
\bm{X}_{v\rightarrow x}\!\!=\! \layernorm(&\softmax({\bm{G}}_{xm})\bm{M}\bm{W}_{xm}\\
&\!\!+\!\softmax(\!\bm{G}_{xf}\!)\bm{F}\bm{W}_{xf}\!+\!\!\bm{X}\bm{W}_x),
\end{aligned}
\end{equation}
\vspace{-5mm}
\begin{equation}
\bm{F}_{x\rightarrow f}= \layernorm(\softmax(\bm{G}_{xf}^T)\bm{X}\bm{W}_{fx}+\bm{F}\bm{W}_f),
\end{equation}
where  $\bm{W}_x$ and $\bm{W}_{fx}$, $\bm{W}_f$, $\bm{W}_{xm}$, and $\bm{W}_{xf}$ are trainable model parameters; the $\softmax(\cdot)$ is applied to the matrix in row-wise; and $\layernorm(\cdot)$~\cite{ba2016layer} denotes layer normalization, which is used for more stable training. Similar to $\bm{F}_{x\rightarrow f}$, we can also compute the influence from hyperedges to clip features $\bm{M}_{x\to m}$. With the cross-modal influences $\bm{X}_{v\rightarrow x}$, $\bm{F}_{x\rightarrow f}$ and $\bm{M}_{x\rightarrow m}$, we can now incorporate them into the original  representations and obtain cross-modality-aware representations, that is,
\begin{align}
\bm{\tilde X} &= \layernorm(\bm{X}_{v\rightarrow x}\bm{W}_{v\rightarrow x} + \bm{X}), \\
{\tilde {\bm{ F}}} &= \layernorm(\bm{F}_{x\rightarrow f}\bm{W}_{x\rightarrow f} + \bm{F}),\\
\bm{\tilde M} &= \layernorm(\bm{M}_{x\rightarrow m}\bm{W}_{x\rightarrow m} + \bm{M}),
\end{align}
where $\bm{W}_{v\rightarrow x}$, $\bm{W}_{x\rightarrow m}$ and $\bm{W}_{x\rightarrow f}$ are trainable model parameters. Actually, $ {\bm{{\tilde{X}}}}$ are the video-aware hyperedge representations, while ${\bm{{\tilde{ F}}}}$ and ${\bm{{\tilde M}}}$ are the question-aware frame and clip representations. Finally, given the video-aware hyperedge representations ${\bm{{\tilde{X}}}}$, we can use it to get the video-aware node (word) representations
\begin{equation} \label{equ:e2n}
{\bm{\tilde Q}} =\bm{D}_v^{-1}\bm{H} {{\bm{\tilde X}}} {{\bm{\tilde W}}},
\end{equation}
where ${\bm{{\tilde{W}}}}$ is a trainable model weights, and $\bm{D}_v$ is the diagonal matrix with diagonal element being the node degree, which is defined as the number of hyperedges connecting to each node. It can be seen that multiplying the incidence matrix $\bm{H}$ in \eqref{equ:e2n} can be viewed as updating the node representation via aggregating information from all the connected hyperedges. The computation above can be seen as a transformation block from $\{\bm{Q}, \bm{F}, \bm{M}\}$ to $\{{\bm{{\tilde Q}}}, {\bm{{\tilde F}}}, {\bm{\tilde{M}}}\}$. Obviously, we can stack more such transformation blocks and constitute a deeper model. 
The influence of depth will be discussed in the experiments.

\subsection{Prediction and Training} \label{sec:Answer Prediction and Loss Function}
Given the cross-modality-aware representations ${\bm{{\tilde{Q}}}}$, ${\bm{{\tilde F}}}$ and ${\bm{{\tilde{M}}}}$, we first project them into a common output space of dimension $d_o$, and then concatenate the projected representations into one matrix $\bm{Y}\in\mathbb{R}^{(N_w+N_f+N_c)\times d_o}$. Afterwards, a self-attention pooling function is applied on $\bm{Y}$ to obtain the  final representation of the entire task $\bm{y}\in\mathbb{R}^{d_o}$ as
\begin{equation}
\bm{y} = \left[\softmax(\LeakyReLU(\bm{Y}\bm{W}_{1}^o)\bm{W}_{2}^o)\right]^T\bm{Y},
\end{equation}
where $\bm{W}_{1}^o\in\mathbb{R}^{d_o\times d_o}$ and $\bm{W}_{2}^o\in\mathbb{R}^{d_o\times 1}$ are trainable model parameters. We design different classifiers for different types of VideoQA tasks, into which the output vector $\bm{y}$ will be fed to predict the answer. For the case of the open-ended format, the output $\bm{y}$ is fed into a linear classifier that outputs the probabilities over the candidates in an answer set $\mathcal{A}$. The classifier is trained by minimizing the cross-entropy loss. It is worth noting that we treat the number counting problem (ranging from 0 to 10) as a regression problem and a $L_2$ regularization is added into the training loss. On the other hand, for the multiple-choice format, we follow previous work HGA~\cite{Jiang2020ReasoningWH} to concatenate the question with every answer candidate from $\mathcal{A}_q$ and generate $N_a$ candidate question-answer sequences for each question. Then, these sequences are fed into our model to produce the output vector $\{\bm{y}_i\}_1^{N_a}$, which are then fed into a linear regression function to output $N_a$ scores for every candidate answer. We train our model by minimizing the hinge loss as
\begin{equation}
\mathcal{L} = \frac{1}{\vert\mathcal{Q}\vert}\sum^{\vert\mathcal{Q}\vert}_{j=1}\sum^{N_a}_{i=1}\max(0, 1+s_i^j-s_t^j),
\end{equation}
where ${\mathcal{Q}}$ is the set of questions; $s_i^j$ denotes the output score for the $i$-th answer of the $j$-{th} question; and $t$ represent the number of ground-truth answer of the $j$-{th} question. In practice, we only need to use a mini-batch from the question set ${\mathcal{Q}}$ for every iteration.

\section{Experiments}
\subsection{Datasets and Baselines}
\noindent{\textbf{Datasets:}} Experiments are conducted on three benchmarks, including TGIF-QA~\cite{Jang2017TGIFQATS}, MSVD-QA~\cite{Xu2017VideoQA}, and MSRVTT-QA~\cite{Xu2017VideoQA} datasets, where the TGIF-QA dataset involves four sub-tasks ($i.e.$, {\em Action}, {\em Transition}, {\em FrameQA} and {\em Count}). 
Details of the three datasets can be found in the Appendix. 
Please note that we use accuracy (Acc.) as the evaluation metric for all the experiments, except repetition count task on TGIF-QA dataset, which uses the Mean Squared Error (MSE).

\noindent{\textbf{Baselines:}} We compare our model with the following strong baselines: AMU~\cite{Xu2017VideoQA}, ST-VQA~\cite{Jang2017TGIFQATS}, Co-Mem~\cite{Gao2018MotionAppearanceCN}, HME~\cite{Fan2019HeterogeneousME}, TSN~\cite{Yang2019QuestionAwareTN}, HGA~\cite{Jiang2020ReasoningWH}, HCRN~\cite{Le2020HierarchicalCR}, L-GCN~\cite{huang2020location}, QueST~\cite{Jiang2020DivideAC}, $Bridge~to~Answer$ (shorted as B2A)~\cite{Park2021BridgeTA}, HAIR~\cite{liu2021hair}, DualVGR~\cite{wang2021dualvgr}.
It is worth mentioning that the performance of baselines on certain datasets are taken from the corresponding papers.

\subsection{Main Results}
The experimental results of our model and the strong baselines on TGIF-QA, MSVD-QA and MSRVTT-QA datasets are shown in Table~\ref{tab:tgif}, Table~\ref{tab:msvd} and Table~\ref{tab:msrvtt}, respectively, with the best performance highlighted in bold. For TGIF-QA dataset, our model outperforms the recent baseline HAIR by 2.6\% on {\em Action}, 2.4\% on {\em Transition}, and 1.3\% on {\em FrameQA}. 
It is worth noting that the B2A baseline also involves syntactic information for VideoQA, but it only takes the whole syntactic graph as a tree, without considering the multi-level compositional semantics of the syntactic information. It can be seen that our model outperforms B2A model by 5.1\%,  2.1\%, and 6.1\% on {\em Action}, {\em Transition}, and {\em FrameQA} sub-tasks, respectively, showing the effectiveness of modeling multi-level compositional semantics of questions. 	The MSVD-QA and MSRVTT-QA benchmarks are more challenging as they only provide open-ended questions. It can be observed that our model also outperforms the most recent DualVGR model targeting on computing question-attended video representation by a large margin (3.3\% and 4.5\% acc. on MSVD-QA and MSRVTT-QA), which shows that semantic composition is essential for representing the semantic meaning of the question.


\subsection{Ablation Studies} \label{sec:Impacts of Different Components}
We conduct in-depth analyses on how different components and parameters contribute to the model performance. 
To this end, we evaluate the performance with different variants of our model from two aspects: (1) excluding or replacing certain components, (2) changing the values of hyperparameters. In addition, we further compare our model with the questions' representation from the pre-trained models.


\begin{table}[!]
	\centering
	\small
	\begin{tabular}{l|cccc}
		\toprule
		Method & Action & Transition & FrameQA & ~$\text{Count}^{\bm{\downarrow}}$~\\\hline
		ST-VQA & 60.8 & 67.1 & 49.3 & 4.40\\
		Co-Mem & 68.2 & 74.3 & 51.5 & 4.10\\
		HME & 73.9 & 77.8 & 53.8 & 4.02\\
		HGA & 75.4 & 81.0 & 55.1 & 4.09\\
		HCRN & 75.0 & 81.4 & 55.9 & 3.82\\
		L-GCN & 74.3 & 81.1 & 56.3 & 3.95\\
		QueST & 75.9 & 81.0 & 59.7 & 4.19\\
		B2A & 75.9 & 82.6 & 57.5 & \textbf{3.71}\\
		HAIR & 77.8 & 82.3 & 60.2 & 3.88\\
		\midrule
		\textit{SCAN} & \textbf{79.8} & \textbf{84.3} & \textbf{61.0} & 3.89\\
		\bottomrule
	\end{tabular}
	\vspace{-2mm}
	\caption{Performances on TGIF-QA dataset.}
	\vspace{-2mm}
	\label{tab:tgif}
\end{table}
\begin{table}[!]
	\centering
	\small
	\resizebox{\linewidth}{!}{
		\begin{tabular}{l|ccccc|c}
			\toprule
			Method & What & Who & How & When & Where & All\\
			\hline
			AMU & 20.6 & 47.5 & 83.5 & 72.4 & 53.6 & 32.0\\
			ST-VQA & 18.1 & 50.0 & \textbf{83.8} & 72.4 & 28.6 & 31.3\\
			Co-Mem & 19.6 & 48.7 & 81.6 & 74.1 & 31.7 & 31.6\\
			HME & 22.4 & 50.1 & 73.0 & 70.7 & 42.9 & 33.7\\
			TSN & 25.0 & 51.3 & \textbf{83.8} & \textbf{78.4} & \textbf{59.1} & 36.7\\
			HGA & 23.5 & 50.4 & 83.0 & 72.4 & 46.4 & 34.7\\
			HCRN & --- & --- & --- & --- & --- & 36.1\\
			QueST & 24.5 & 52.9 & 79.1 & 72.4 & 50.0 & 36.1\\
			B2A & --- & --- & --- & --- & --- & 37.2\\
			HAIR & --- & --- & --- & --- & --- & 37.5\\
			DualVGR & 28.7 & 53.8 & 80.0 & 70.7 & 46.4 & 39.0\\
			\midrule
			\textit{SCAN} & \textbf{29.5} & \textbf{55.7} & 82.4 & 72.4 & 42.9 & \textbf{40.3}\\
			\bottomrule
		\end{tabular}
	}
	\vspace{-2mm}
	\caption{Performances on MSVD-QA dataset.}
	\vspace{-2mm}
	\label{tab:msvd}
\end{table}
\begin{table}[!h]
	\centering
	\small
	\resizebox{\linewidth}{!}{
		\begin{tabular}{l|ccccc|c}
			\toprule
			Method & What & Who & How & When & Where & All\\
			\hline
			AMU & 26.2 & 43.0 & 82.4 & 72.5 & 30.0 & 32.5\\
			ST-VQA & 24.5 & 41.2 & 78.0 & 76.5 & 34.9 & 30.9\\
			Co-Mem & 23.9 & 42.5 & 74.1 & 69.0 & 42.9 & 32.0\\
			HME & 26.5 & 43.6 & 82.4 & 76.0 & 28.6 & 33.0\\
			TSN & 27.9 & 46.1 & \textbf{84.1} & 77.8 & \textbf{37.6} & 35.4\\
			HGA & 29.2 & 45.7 & 83.5 & 75.2 & 34.0 & 35.5\\
			HCRN & --- & --- & --- & --- & --- & 35.6\\
			QueST & 27.9 & 45.6 & 83.0 & 75.7 & 31.6 & 34.6\\
			B2A & --- & --- & --- & --- & --- & 36.9\\
			HAIR & --- & --- & --- & --- & --- & 36.9\\
			DualVGR & 29.4 & 45.6 & 79.8 & 76.7 & 36.4 & 35.5\\
			\midrule
			\textit{SCAN} & \textbf{30.3} & \textbf{48.8} & 81.5 & \textbf{78.0} & 37.2 & \textbf{37.1}\\
			\bottomrule
		\end{tabular}
	}
	\vspace{-2mm}
	\caption{Performances on MSRVTT-QA dataset.}
	\label{tab:msrvtt}
	\vspace{-3.5mm}
\end{table}

\paragraph{Impacts of Different Components}

\begin{figure}[!t]
	\centering
	\includegraphics[scale=0.4]{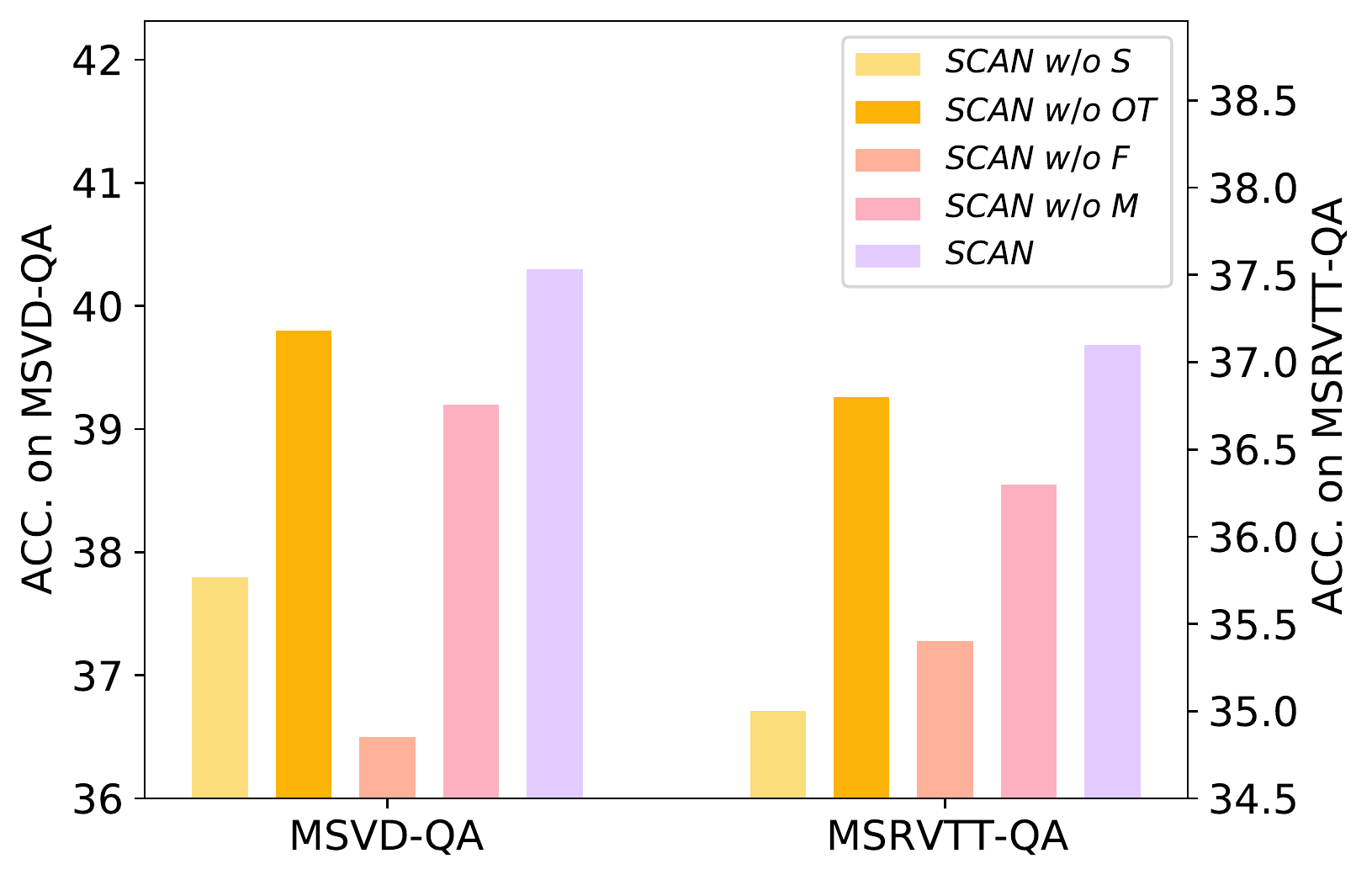}
	\vspace{-2mm}
	\caption{Performances of \textit{SCAN} variants that exclude or replace certain components on MSVD-QA and MSRVTT-QA datasets.}
	\label{fig:acc_bar}
	\vspace{-2mm}
\end{figure}

We evaluate the effectiveness of different components by eliminating four modules. \emph{1)~Syntactic tree (\textit{SCAN} w/o S)}: we remove the syntactic hypergraph and only consider the word-level cross-modality alignment. \emph{2)~Optimal transport alignment (\textit{SCAN} w/o OT)}: we remove the optimal transport alignment module and replace it with a simple dot-product similarity module. \emph{3)~Frame-level feature (\textit{SCAN} w/o F)}: we remove the frame-level video feature and only keep the motion-level feature. \emph{4)~Clip-level feature (\textit{SCAN} w/o M)}: we remove the clip-level video feature. We compare these four tailored models with the complete \textit{SCAN} model on the MSVD-QA and MSRVTT-QA datasets. The results are showed in Fig.~\ref{fig:acc_bar}. It can be seen that eliminating syntactic hypergraph \emph{(\textit{SCAN} w/o S)} leads to significant performance drop, suggesting the importance of modeling compositional semantics of the question. It can be further observed that OT alignment \emph{(\textit{SCAN} w/o OT)} also largely contributes to the model performance, demonstrating the necessity of filtering out irrelevant video information by using a more sparser alignment matrix. Also, eliminating frame-level and clip-level features, $i.e.$, \emph{\textit{SCAN} (w/o F)} and \emph{\textit{SCAN} (w/o M)}, also harms the final performance, which indicates the importance of simultaneously modeling visual features at different levels.

\begin{figure}[!t]
	\centering
	\subfigure[Action]{
		\begin{minipage}[t]{0.3\linewidth}
			\centering
			\includegraphics[width=1.0in]{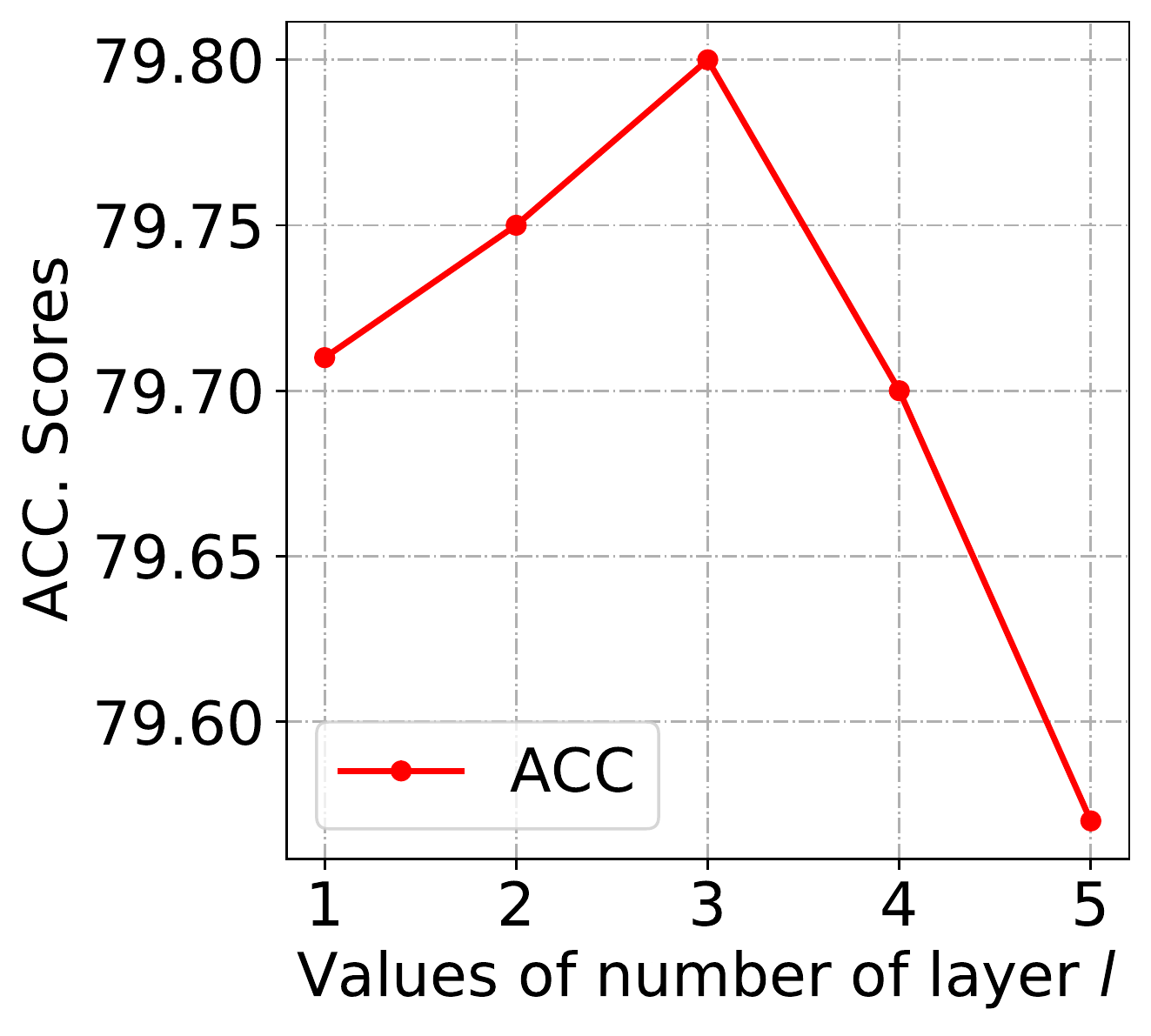}
		\end{minipage}
	}%
	\subfigure[Transition]{
		\begin{minipage}[t]{0.3\linewidth}
			\centering
			\includegraphics[width=0.97in]{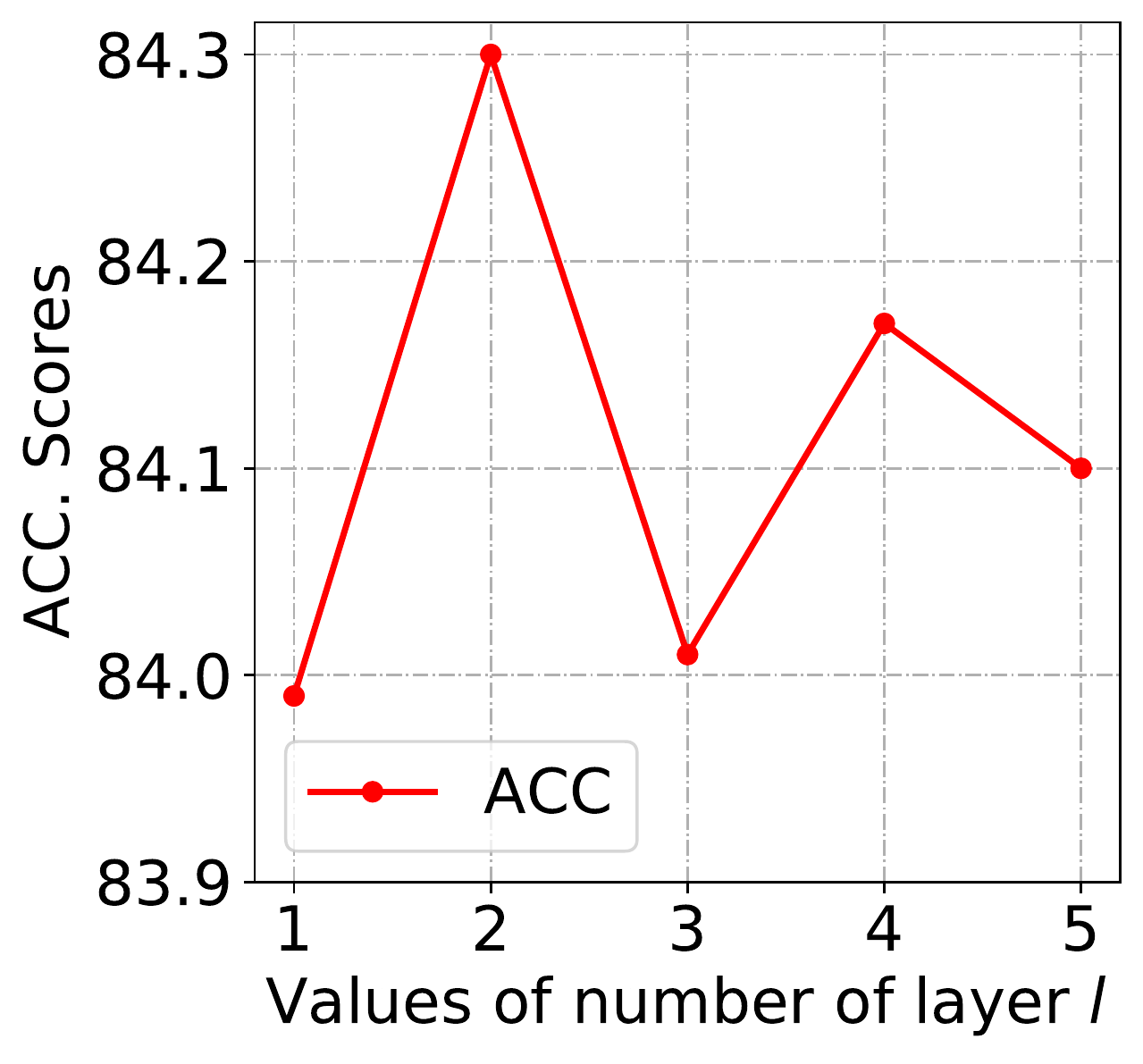}
		\end{minipage}%
	}%
	\subfigure[FrameQA]{
		\begin{minipage}[t]{0.3\linewidth}
			\centering
			\includegraphics[width=0.97in]{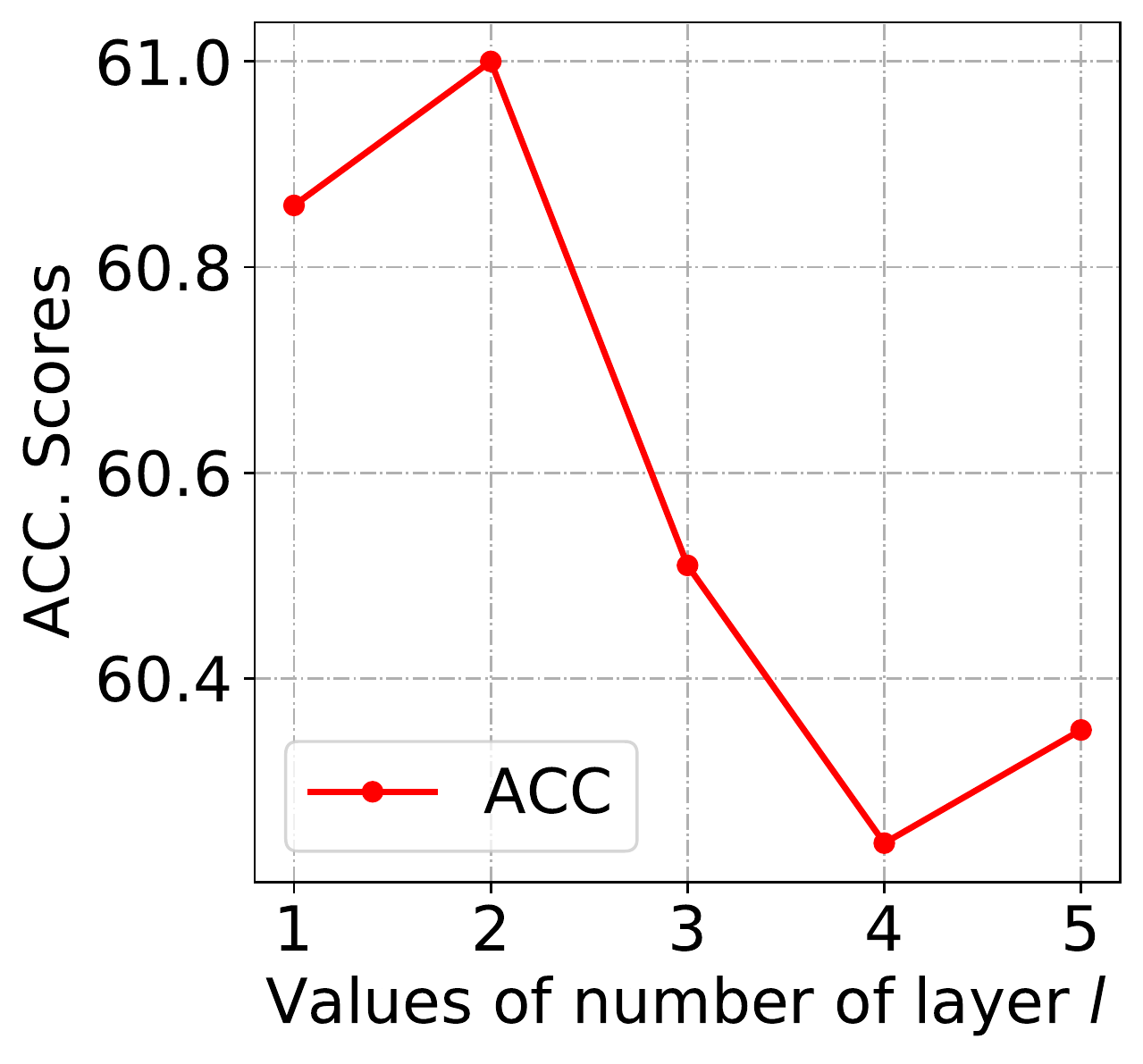}
		\end{minipage}
	}%
	\vspace{-3mm}
	\caption{The performance of \textit{SCAN} under different number $l$ of the video-aware hypergraph convolutional network layers on $Action$, $Transition$, and $FrameQA$ tasks.}
	\label{fig:acc_line}
	\vspace{-3mm}
\end{figure}

\begin{table}[!t]
	\small
	\tabcolsep=0.32cm
	\centering
	\begin{tabular}{c|c|c}
		\hline
		& MSVD-QA & MSRVTT-QA \\\hline
		\hline
		$\text{BERT}_{large}$-fixed & 37.94 & 35.16 \\\hline
		$\text{BERT}_{large}$ & 38.70 & 36.12 \\\hline
		\hline
		\textit{SCAN} & \textbf{40.3} & \textbf{37.1} \\\hline
	\end{tabular}
	\caption{Performances of models that learn representation based on syntactic hypergraph or with pre-trained BERT model on MSVD-QA and MSRVTT-QA datasets.}
	\vspace{-6mm}
	\label{tab:acc_bert}
\end{table}

\begin{figure*}[!t]
	\centering
	\subfigure[Word-level: \textit{net}]{
		\begin{minipage}[t]{\linewidth}
			\centering
			\includegraphics[width=5.51in]{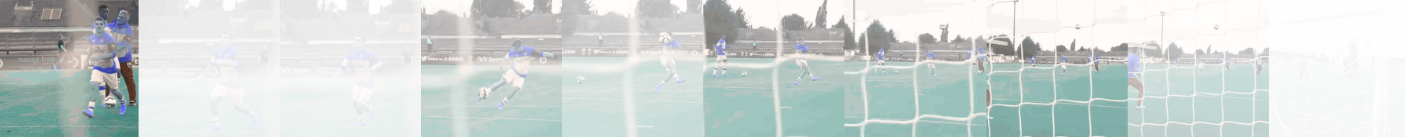}
		\end{minipage}
	}%
	\vspace{-1mm}
	\subfigure[Phrase-level: \textit{gets into the net}]{ 
		\begin{minipage}[t]{\linewidth}
			\centering
			\includegraphics[width=5.51in]{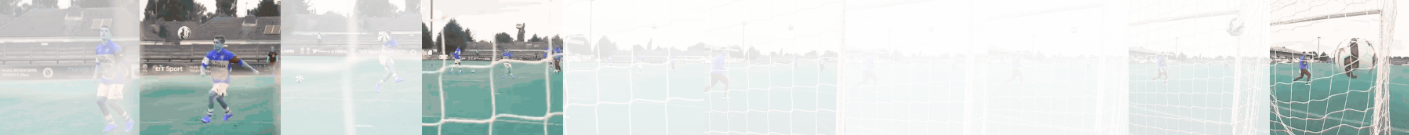}
		\end{minipage}%
	}%
	\vspace{-1mm}
	\subfigure[Sentence-level: \textit{what is the soccer player wearing a uniform gets into the net}]{
		\begin{minipage}[t]{\linewidth}
			\centering
			\includegraphics[width=5.51in]{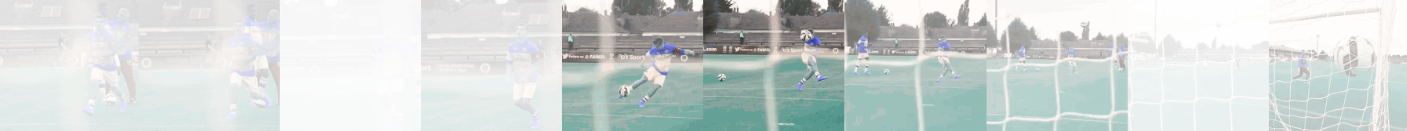}
		\end{minipage}%
	}%
	\vspace{-1mm}
	\caption{Visualization of visual-language alignments between semantic composition and video frames.}
	\vspace{-5mm}
	\label{fig:case1}
\end{figure*}

\begin{figure}[!t]
	\centering
	\subfigure[Simple Dot-production]{
		\begin{minipage}[t]{0.49\linewidth}
			\centering
			\includegraphics[width=1.3in]{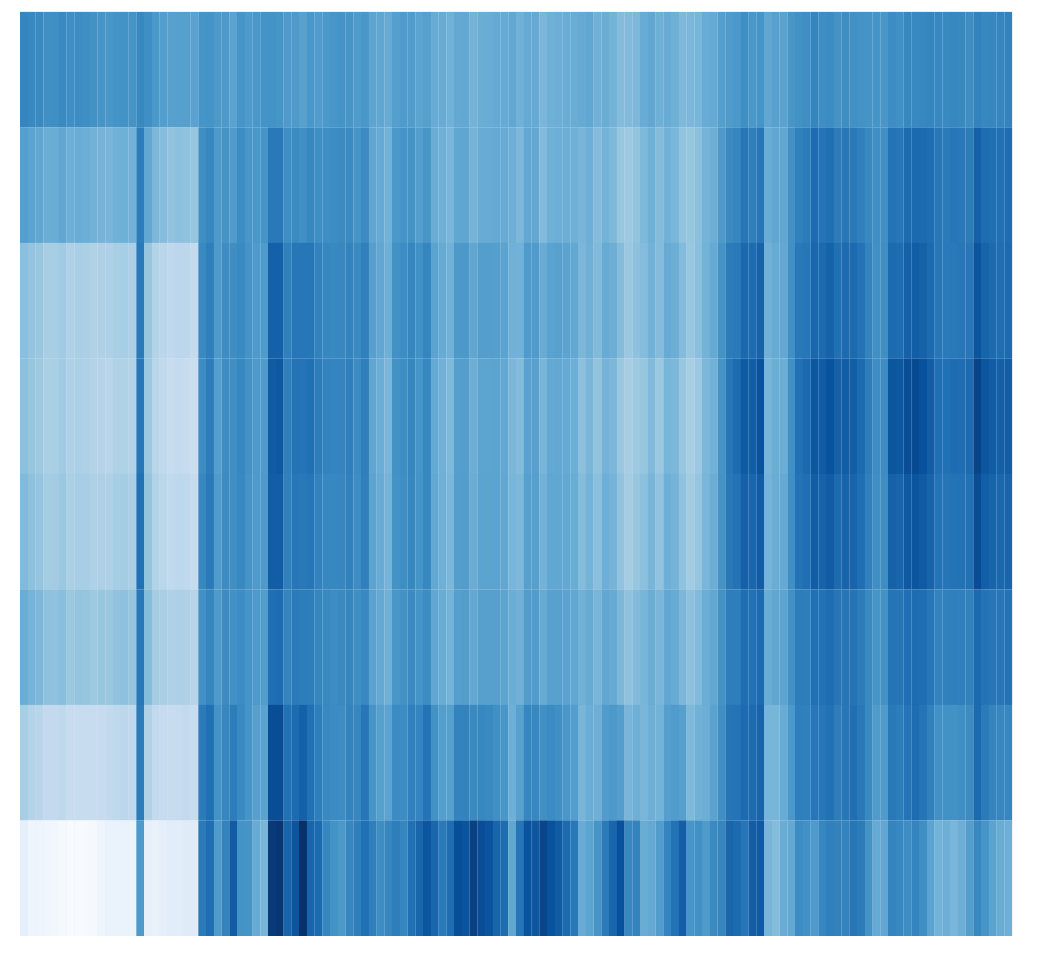} 
		\end{minipage}
	}%
	\subfigure[Optimal Transport]{
		\begin{minipage}[t]{0.49\linewidth}
			\centering
			\includegraphics[width=1.55in]{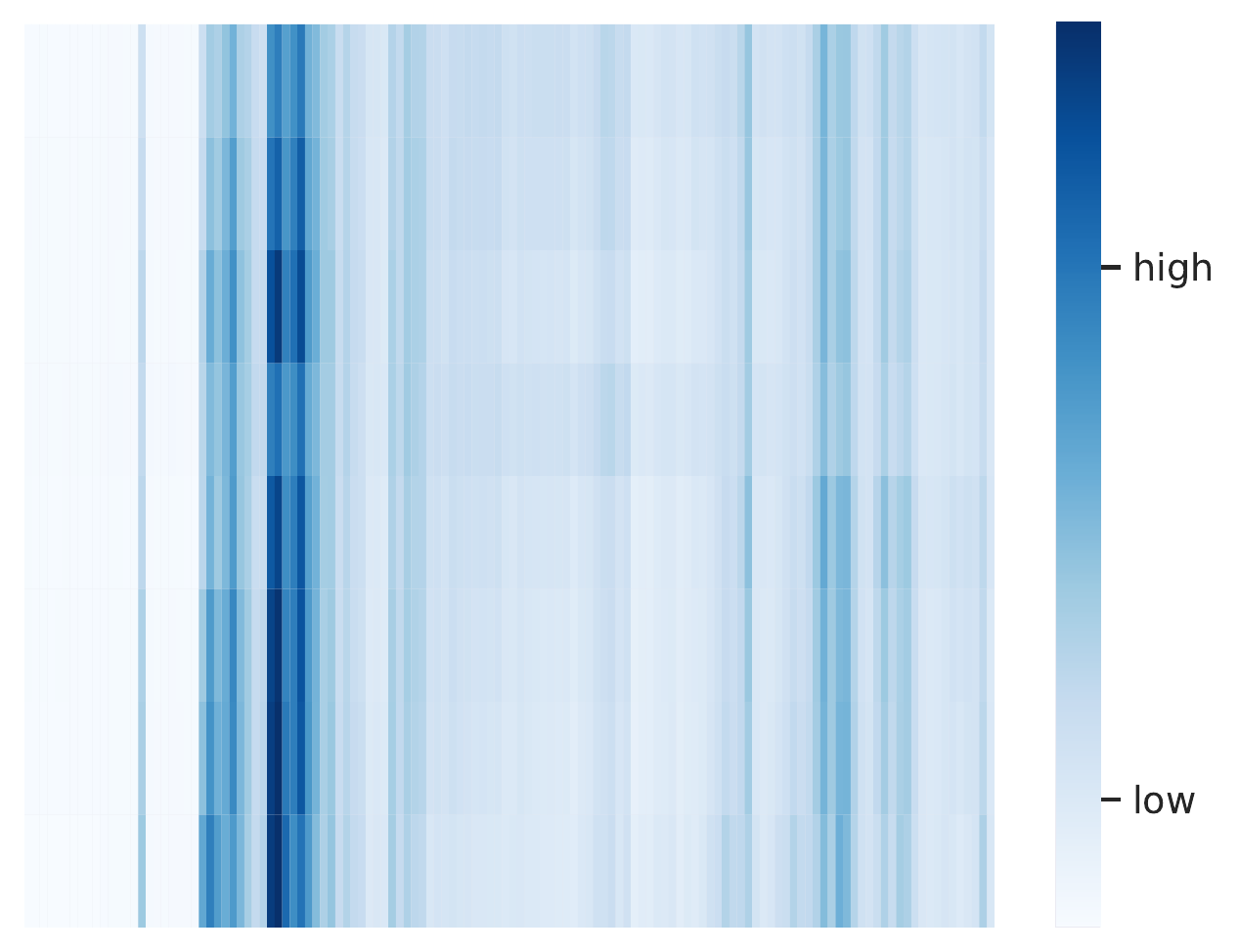} 
		\end{minipage}%
	}%
	\vspace{-2mm}
	\caption{Visualization of different attention matrices.}
	\vspace{-3mm}
	\label{fig:case2}
\end{figure}

\paragraph{Impacts of the Number of Computational Blocks} 
We analyze the impact of using different number of computation blocks in the proposed HCN.
Fig.~\ref{fig:acc_line} shows the performance on {\em Action}, {\em Transition} and {\em FrameQA} sub-tasks of TGIF-QA dataset with the number changing from 1 to 5. It can be observed that the optimal number varies from task to task, {\it e.g.}, $l=2$ for {\em Transition} and {\em Frame QA}, and $l=3$ for the {\em Action} subtask. This phenomenon indicates that different question types emphasize different levels of compositional semantics of questions, and we can increase the depth of the interaction layers to support more complex task, or find a tradeoff between performance and efficiency.

\vspace{-2mm}
\paragraph{Comparison with the Pre-trained Models} 
To better analyze how syntactic hypergraph contributes to the model's performance, we further conduct the ablation study that excludes the entire syntactic hypergraph and adopts the pre-trained model BERT~\cite{devlin-etal-2019-bert} to learn the questions' representation. It can be seen from Table~\ref{tab:acc_bert} that, our model outperforms the $\text{BERT}_{large}$ model. Although some implicit relationships between words can be captured by the pre-trained models~\cite{xu-etal-2021-syntax}, the experimental results once again confirm the effectiveness of explicitly exploring the syntactic composition via our proposed syntactic hypergraph.

\subsection{Qualitative Analysis} \label{sec:Case Study}

\paragraph{Visualization of Language-Vision Alignments} 
In order to visualize the alignments between different semantic composition and the visual feature, we take the question ``\textit{what is the soccer player wearing a uniform gets into the net}'' extracted from $frameQA$ as an example.
We visualize the aligned videos based on different levels of semantic composition (i.e., word-level, phrase-level and sentence-level).
Then we extract 10 frames, which in temporally ranked, with the highest alignment weights for each semantic level.
The visualization is shown in Fig.~\ref{fig:case1}, where frames with higher attention weights are clearer and vice verse. 
It can be seen that, the word $net$ is aligned with most frames containing nets. 
When it is composed to phrase-level semantics (``\textit{gets into the net}"), the matched frames intend to focus on the scene that the soccer goes into the net.
Finally, for sentence-level, we find that key frames focus on the entire process of a football player kicking the ball into the net.
The results show that \textit{SCAN} can better model the semantic composition phenomenon and its multi-modal alignment with the visual information.

\paragraph{Visualization of Different Alignment Matrices} To analyze whether our OT Alignment mechanism indeed generate sparse alignment scores, we visualize the OT alignment matrix and compare it with the matrix generated by simply dot-product approach, and plots one typical example in Fig.~\ref{fig:case2}. It can be seen that the OT method returns a sparser alignment matrix, while dot-product-based attention is effectively dense. This emphasizes the effectiveness of OT-based soft alignment in concentrating more relevant cross-modal information.

\section{Conclusion}
In this paper, we propose to model the semantic composition phenomenon of question with a syntactic hypergraph for VideoQA. We first build the syntactic hypergraph based on the syntactic dependency tree in a hierarchical bottom-up manner. Then we propose a cross-modality-aware syntactic hypergraph convolution network to align the cross-modal semantic information. To enhance the cross-modal alignment, we adopt the optimal transport attention mechanism to obtain a sparse matching. Experiments show that our method outperforms strong baselines on three benchmark datasets and verify the effectiveness of each component.




\bibliography{anthology,custom}
\bibliographystyle{acl_natbib}

\appendix
\section{More Details for the Datasets}
Experiments are conducted on three benchmark datasets, including TGIF-QA~\cite{Jang2017TGIFQATS}, MSVD-QA~\cite{Xu2017VideoQA}, and MSRVTT-QA~\cite{Xu2017VideoQA} datasets.

\emph{1)~TGIF-QA}~\cite{Jang2017TGIFQATS} is a prominent large-scale benchmark dataset for VideoQA task that consists of 165$K$ Q\&A pairs based on 72$K$ animated GIFs. The dataset defines four tasks: (1) Repeating action ($Action$) requires to identify the action repeated for a given number of times from 5 candidate answers; (2) State transition ($Transition$) also deals with 5 candidate answers aiming to identify the transition of two states; (3) Frame QA ($FrameQA$) is an open-ended task that needs to find a key frame in the video to indicate the correct answer from a pre-defined dictionary; (4) Repetition count ($Count$) also contains an open-ended numbers of task to count the number of occurrences of an action.

\emph{2)~MSVD-QA}~\cite{Xu2017VideoQA} is an open-ended VideoQA dataset, which is divided into 5 different types, including $what$, $who$, $how$, $when$, and $where$. The dataset contains 1970 short video clip, 50505 Q\&A pairs and 1000 pre-defined answers.

\emph{3)~MSRVTT-QA}~\cite{Xu2017VideoQA} is similar to MSVD-QA. It is also divided into the same five types with 1000 pre-defined answers. The MSRVTT-QA is generated from the MSRVTT~\cite{Xu2016MSRVTTAL} dataset, containing 10K videos and 243$K$ Q\&A.

\section{Algorithm for OT} \label{app:Algorithm for OT}
We adopt an off-the-shelf differentiatable approximate method~\cite{xie2020fast} to obtain the OT matrix $\bm{\pi}^*$, which is summarized in Algorithm~\ref{alg:ot}.

\begin{algorithm*}[!t]
	\caption{OT-based Attention Mechanism $\mathcal{OT}(\bm{X},\bm{F})$}
	\label{alg:ot}
	\begin{algorithmic}[1] 
		\State \textbf{Input}: Hyperedges matrices $\bm{X}=\{\bm{x}_i\}_1^{N_s}$ frame-level matrices $\bm{F}=\{\bm{f}_j\}_1^{N_f}$.
		\State $\bm{b}=\frac{1}{N_s}\bm{1}_{N_s}$, $\bm{\pi}^{(1)}=\bm{1}_{N_s}\bm{1}_{N_f}^T$, $\bm{C}_{ij}=e^{-c(\bm{x}_i,\bm{f}_j)}$.
		\For{$t=1,2,\cdots,10$}
		\State $\bm{\Gamma}=\bm{C}\odot\bm{\pi}^{(t)}$.
		\State $\bm{a}=\frac{1}{N_s\bm{\Gamma}\bm{b}}$, $\bm{b}=\frac{1}{N_f\bm{\Gamma}^T\bm{a}}$.
		\State $\bm{\pi}^{(t+1)}=\text{diag}(\bm{a})\bm{\Gamma}\text{diag}(\bm{b})$.
		\EndFor
		\State \textbf{Return} $\bm{\pi}$
	\end{algorithmic}
\end{algorithm*}
\begin{algorithm*}[!t]
	\caption{Subtree Generation Algorithm SubTreeGen($\cdot,\cdot$)}
	\label{alg:subtree}
	\begin{algorithmic}[1] 
		\State \textbf{Input}: Syntactic dependency Tree $\mathcal{T}$, set of vertices $\mathcal{V}$ containing all vertices on the tree.
		\State $\mathcal{T}_s=\{\}$ // The set containing all found subtrees.
		\For{$v_i$ in $\mathcal{V}$}
		\If{$v_i$ is leaf node}
		\State \textit{APPEND}($\mathcal{T}_s$, $\{v_i\}$)
		\EndIf
		\If{$v_i$ is branch node}
		\For{$c$ in Child($\mathcal{T}, v_i$)} // The child of $v_i$
		\State \textit{APPEND}($\mathcal{T}_s$, $\{v_i, \text{GetSubTree}(\mathcal{T},c)\}$)
		\EndFor
		\State \textit{APPEND}($\mathcal{T}_s$, $\{\text{GetSubTree}(\mathcal{T},v_i)\}$)
		\EndIf
		\EndFor
		\State \textbf{Return} $\mathcal{T}_s$
	\end{algorithmic}
\end{algorithm*}
\vspace{-3mm}
\begin{algorithm*}
	\caption{GetSubTree($\mathcal{T}$,$c$)}
	\begin{algorithmic}[1]
		\State \textbf{Input}: Syntactic dependency Tree $\mathcal{T}$, node $c$.
		\If{$c$ is leaf node}
		\State \textbf{Return} $\{c\}$
		\EndIf
		\State $\mathcal{T}'_s=\{c\}$
		\For{$c'$ in Child($\mathcal{T}, c$)}
		\State \textit{APPEND}($\mathcal{T}'_s$, GetSubTree($\mathcal{T}$,$c'$))
		\EndFor
		\State \textbf{Return} $\mathcal{T}'_s$
	\end{algorithmic}
\end{algorithm*}

\section{Subtree Generation Algorithm}
Algorithm~\ref{alg:subtree} shows our subtree generation algorithm SubTreeGen$(\cdot,\cdot)$ used in our syntactic hypergraph construction. The algorithm begins by taking each leaf node as a subtree. Then, for the branch node $v_i$, our algorithm first adopts the recursive GetSubRree$(\cdot,\cdot)$ on each of the child node of node $v_i$ to obtain the connected subtrees. Secondly, we add node $v_i$ to all connected subtrees to generate more trees. Finally, we directly apply recursive GetSubRree$(\cdot,\cdot)$ on node $v_i$ to obtain higher level semantic composition.

\end{document}